\newcommand{\remove}[1] {}

\newcommand{\parskiny}{\vspace{-3mm}}
\newcommand{\seckiny}{\vspace{-2mm}}
\newcommand{\figskiny}{\vspace{-4mm}}
\newcommand{\figshrinky}{\vspace{-2mm}}

\documentclass[10pt,twocolumn,letterpaper]{article}

\usepackage{cvpr}
\usepackage{times}
\usepackage{graphicx}
\usepackage{amsmath,amssymb} 
\usepackage{color}

\cvprfinalcopy

\usepackage{epsfig}
\usepackage{floatrow}
\usepackage{array}
\usepackage{slashbox}
\usepackage{multirow}
\usepackage{caption}
\usepackage{subcaption}
\usepackage[numbers,sort&compress,square,comma]{natbib}
\usepackage[pagebackref=false,breaklinks=true,letterpaper=true,colorlinks,bookmarks=false]{hyperref}

\newcommand{\oldtext}[1]{} 

\DeclareMathAlphabet{\mathpzc}{T1}{pzc}{m}{n}

\def\cvprPaperID{828} 

\ifcvprfinal\fi

\begin{document}

\title{Context Forest for efficient object detection with large mixture models}

\author{Davide Modolo\\
University of Edinburgh\\
{\tt\small d.modolo@sms.ed.ac.uk}
\and
Alexander Vezhnevets\\
University of Edinburgh\\
{\tt\small avezhnev@inf.ed.ac.uk}
\and
Vittorio Ferrari\\
University of Edinburgh\\
{\tt\small vferrari@staffmail.ed.ac.uk}
}

\maketitle

\begin{abstract}

We present Context Forest (ConF) --- a technique for predicting properties of the objects in an image based on its global appearance. Compared to standard nearest-neighbour techniques, ConF is more accurate, fast and memory efficient. 
We train ConF to predict which aspects of an object class are likely to appear in a given image (e.g. which viewpoint).
This enables to speed-up multi-component object detectors,
by automatically selecting the most relevant components to run on that image.
This is particularly useful for detectors trained from large datasets, which typically need many components to fully absorb the data and reach their peak performance.
ConF provides a speed-up of 2x for the DPM detector~\cite{felzenszwalb10pami} and of 10x for the EE-SVM detector~\cite{MalisiewiczICCV11}. 
To show ConF's generality, we also train it to predict at which locations objects are likely to appear in an image.
Incorporating this information in the detector score improves mAP performance by about 2\% by removing false positive detections in unlikely locations.
\vspace{-2mm}


\end{abstract}

\seckiny
\section{Introduction}
\label{sec:intro}

\begin{figure*}[t]
\includegraphics[width=\textwidth]{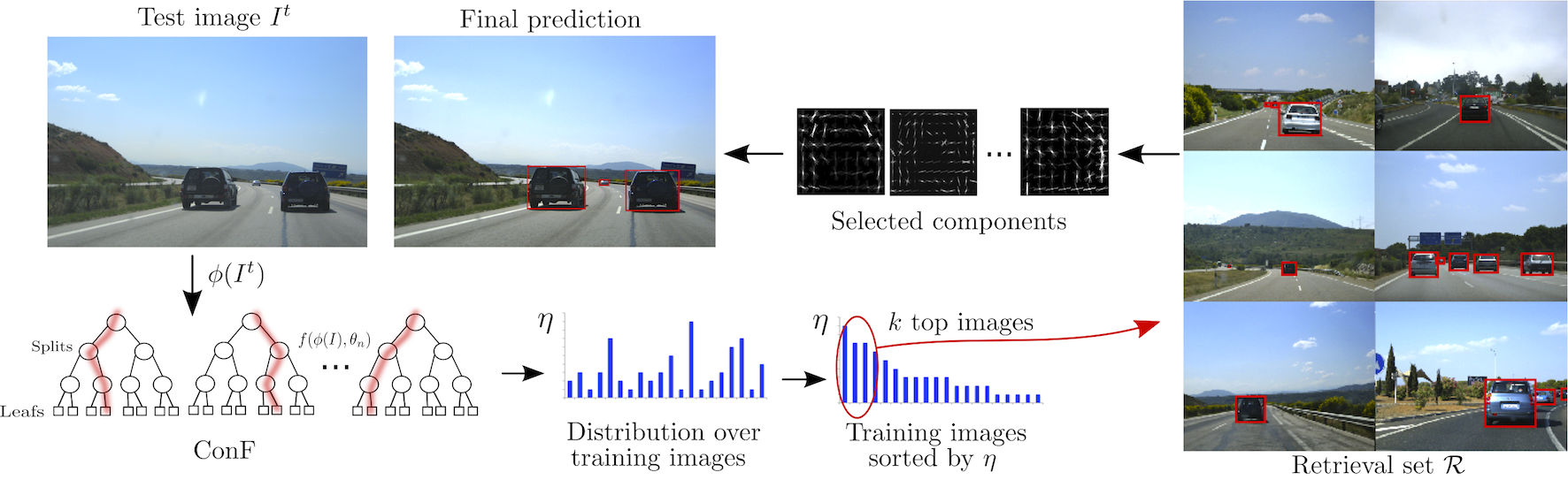}
\caption{\small \it Illustration of ConF selecting components for a test image (sec.~\ref{sec:acs}.) \figskiny } 
\label{fig:highway}
\end{figure*}

\begin{figure*}[t]
\includegraphics[width=\textwidth]{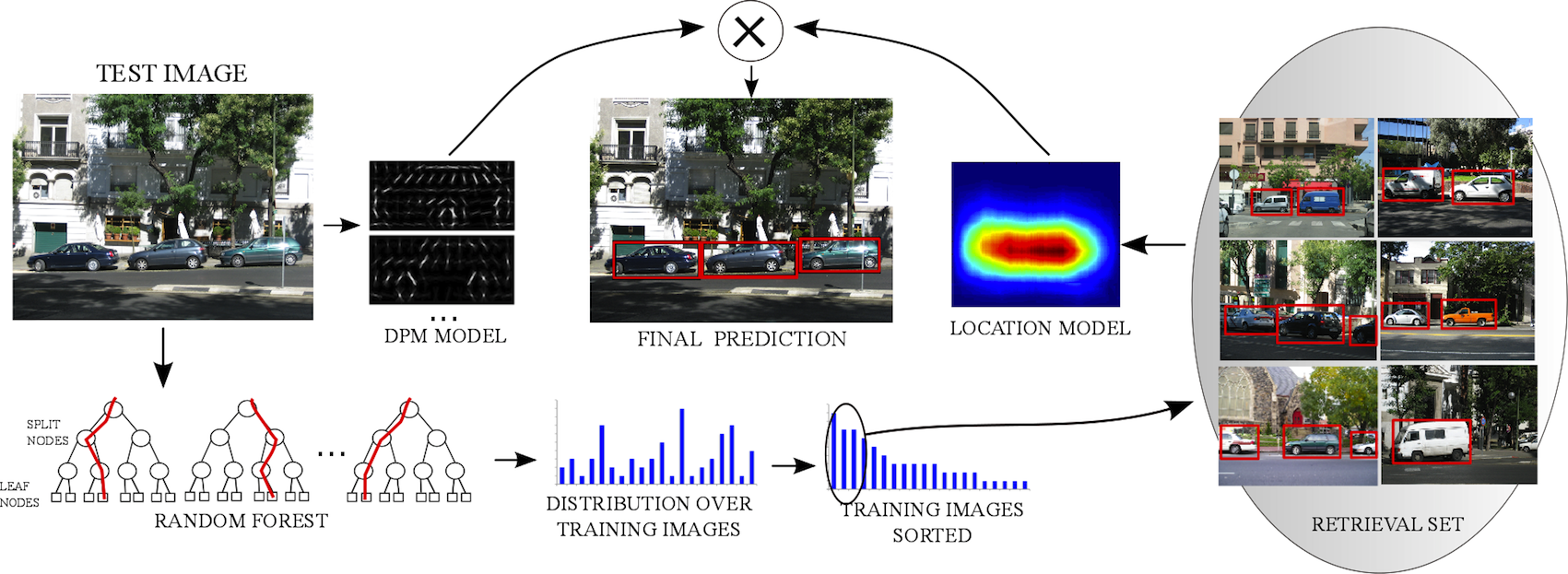}
\caption{\it \small Schematic illustration ConF predicting object locations and their use to improve the DPM score (sec.~\ref{sec:loc}). \figskiny} 
\label{fig:schema}
\end{figure*}

Global image appearance carries information about properties of objects in the image, such as their appearance and location. For instance, a picture of a highway taken from a car is more likely to contain cars from the back viewpoint than from the side (fig.~\ref{fig:highway}). A picture of a racing track is more likely to contain racing cars than minivans. This also applies to other classes. A person in a road scene is more likely to be standing than sitting. Another property that can be inferred from global image appearance is the rough location of object instances~\cite{russell07nips}. For instance, an urban scene with cars parked in front of a building, shows cars in the bottom half of the image (fig.~\ref{fig:schema}).

In this paper we exploit this observation for the benefit of object detection. We propose a method, coined Context Forest (ConF), for learning the relation between the global image appearance and the properties of the objects it contains.
Given only the global appearance of a test image, ConF retrieves a subset of training images that contain objects with similar properties.
ConF is based on the Random Forest~\cite{Breiman01,criminisi2011} framework, which provides high computational efficiency and the ability to learn complex, non-linear relations between global image appearance and objects properties. It is very flexible and only requires these properties to be defined through a distance function between two object instances, e.g. their appearance similarity or difference in location.
We demonstrate ConF by learning to predict two properties: aspects of objects appearance and location.
ConF trained to predict appearance is then used to speed up multi-component object detectors~\cite{felzenszwalb10pami,MalisiewiczICCV11} and ConF trained for object location is used to remove false positives.

Multi-component detectors~\cite{felzenszwalb10pami,MalisiewiczICCV11} model the appearance variations within an object class as a mixture of several components. Each component is trained to recognize a particular aspect of objects appearance. For example, cars could have viewpoint components~\cite{felzenszwalb10pami}, such as front and back views, or subclass components~\cite{divvala2012eccv}, such as taxi, ambulance and minivan. 
There is growing evidence~\cite{zhu12bmvc} that the performance of object detectors tends to saturate as the amount of training data increases.
We conduct an extensive experiment (sec.~\ref{sec:large_scale_MC}) on a dataset containing 15x the amount of training data than PASCAL VOC 2012~\cite{pascal-voc-2012} using two popular multi-component detectors: Deformable Part-based Model~\cite{felzenszwalb10pami} (DPM) and the Ensemble of Exemplar SVMs~\cite{MalisiewiczICCV11} (EE-SVM). 
Our results show that as the size of training set grows, these detectors can absorb the additional intra-class appearance variations and continue to improve their performance, but only if the amount of components is increased accordingly. This extends the related findings of~\cite{zhu12bmvc} on single linear SVM HOGs to the DPM and EE-SVM cases.
Although increasing the number of components significantly improves performance, it also makes the detectors much slower, as all components need to be run on every test image. 

We use ConF to select a subset of model components which is most relevant to a particular test image. We then run only those components, obtaining a speed-up. Our experiments show that ConF delivers a 2x speed-up for DPM~\cite{felzenszwalb10pami} and 10x speed-up for EE-SVM~\cite{MalisiewiczICCV11} without loss of accuracy (sec.~\ref{sec:exp}).
Hence, ConF makes large multi-component detectors practical. This is particularly useful for EE-SVMs, as their number of components is truly very large (i.e. as many as there are training instances).
Interestingly, in some cases we even gain a small improvement in accuracy, by not running some components that would produce false positive detections.

Moreover, we train a second ConF to predict at which positions and scales objects are likely to appear in a given test image, analogue to~\cite{russell07nips}. By incorporating this information in the detector score at test time, we reduce the false positive rate by removing detections in unlikely locations. Experiments show an mAP improvement of 2\%. This demonstrates that ConF is a general technique that can predict various kinds of object properties.

Finally, we carry out an extensive comparison to standard nearest-neighbour techniques for such context-based predictions~\cite{liu09cvpr,russell07nips,Torralba03,rabinovich07iccv,tighe10eccv}, which shows that ConF predicts object properties from global image appearance more accurately, it is much faster and more memory efficient (sec.~\ref{sec:exp}).

The rest of the paper is organized as follows. We start by reviewing related work in sec.~\ref{sec:related_work}. Sec.~\ref{sec:forest} explains ConF, our main contribution. In sec.~\ref{sec:large_scale_MC} we present a first series of experiments, which study the behaviour of multi-component detectors on large training sets and thus motivate ConF for component selection. Finally, we present a second series of experiments to validate the benefits of ConF on two object detectors in sec.~\ref{sec:exp}.




\section{Related work} 
\label{sec:related_work}




\paragraph{Context.}
The use of context for object detection is a broad research area. Some works~\cite{rabinovich07iccv,Heitz08,choi10cvpr,desai:iccv09} 
model context as the interactions between multiple object classes in the same image. In this paper, we model context as a relation between global image appearance and properties of the objects within them, as in~\cite{liu09cvpr,russell07nips,Torralba03,tighe10eccv}. 
These works have shown that global image descriptors give a valuable cue about which classes might be present in an image and where they are located. Since then, many object detectors~\cite{murphy03nips,felzenszwalb10pami,harzallah2009iccv,uijlings13ijcv,ilsvrc13euuva} employed such global context to re-score their detections, thereby removing out-of-context false-positives.
A similar approach was taken by~\cite{liu09cvpr,tighe10eccv} for image parsing.
All of these works have a nearest neighbour core:
they first retrieve a small subset of training images which are most globally similar to a test image,
and then transfer the relevant statistics of the object properties in this retrieval set to the test image.
In our work instead the retrieval set is estimated by ConF, which is {\em explicitly trained} to return images containing objects with similar properties to those in the test image.
ConF has several advantages over nearest-neighbour approaches:
(i) it can learn highly complex non-linear dependencies between the global descriptor and the object property. As a result, it estimates it more accurately;
(ii) in large training sets, nearest neighbour becomes very slow, as its complexity is linear in their size. ConF is much faster and more memory efficient;
(iii) ConF supports any objective function, which might even be evaluated on a different data representation than the input at test time. This is a crucial feature for our problem, as we want to predict properties of objects, but based on global image features.

\paragraph{Multi-component detectors.}
These detectors~\cite{felzenszwalb10pami, gu12eccv, drayer14eccv, divvala2012eccv, aghazadeh2012eccv, MalisiewiczICCV11} model each aspect of an object class as a separate component. They are very popular but can be slow when trained from large training sets as they need many components to reach peak performance.
While we present experiments on DPM~\cite{felzenszwalb10pami} and EE-SVM~\cite{MalisiewiczICCV11}, ConF can benefit all kind of multi-component detectors, and it allows them to use large training sets without compromising speed.

\paragraph{EE-SVM.}
The EE-SVM~\cite{MalisiewiczICCV11} is an extreme case of multi-component detector, where a separate component is created for each training example. Due to that EE-SVM benefit the most from dynamically selecting components with ConF.
EE-SVMs are widely used for many applications beyond object detection~\cite{singh12eccv, aytar12bmvc, endres13cvpr, dong13cvpr, tighe13cvpr, Gu10eccv,jain11cvpr}.
As they explicitly associate a training example to an object in the test image, they enable transferring meta-data such as segmentation masks~\cite{tighe13cvpr,MalisiewiczICCV11}, 3D models~\cite{MalisiewiczICCV11}, subcategory~\cite{jain11cvpr} and viewpoint~\cite{Gu10eccv}.
They can as well be used for discovering mid-level discriminative patches~\cite{endres13cvpr, singh12eccv, juneja13cvpr, doersch12siggraph}
for scene classification~\cite{juneja13cvpr,singh12eccv},
for forming a dictionary of object parts~\cite{endres13cvpr, singh12eccv},
or to characterize a certain geo-spatial area~\cite{doersch12siggraph},
All these applications can potentially be sped-up by ConF. 




\section{Estimating object properties from context}
\label{sec:forest}



In this section we exploit the observation that global image appearance contains information about properties of the objects inside it. We focus on two kinds of properties: aspects of appearance and location in the image. We propose a new method, coined Context Forest (ConF), based on the Random Forest framework~\cite{Breiman01,criminisi2011}, which learns the relation between global image features and the properties of the object in that image. Given only the global image appearance of a test image, ConF retrieves a subset of training images that contain objects with similar properties. 
\remove{By training ConF to retrieve images with similar object appearance we can select which model components are most relevant for a given test image. Running only the selected components leads to a speed-up. Interestingly, this can even improve detection performance, as it removes some false positives produced by components not relevant to the contents of this particular test image (sec.~\ref{sec:exp}). 
Moreover, we can utilize ConF to predict at which position and scale objects are likely to appear in that test image. This yields an estimate of the probability that a test window contains an object instance, based purely on its location in the image. We then augment the detector's score with this probability, which improves performance as it downgrades the score of false positives at unlikely locations.} 


\seckiny
\subsection{Context Forest (ConF)}
\label{sec:acs}

Given a training set $\mathcal{T}$ the goal of ConF is to map the global appearance $\phi(I_t)$ of a test image $I_t$ into a retrieval set $\mathcal{R} \subset \mathcal{T}$. We want to construct a mapping, such that properties of objects (e.g. appearance, location) in images of $\mathcal{R}$ are similar to the properties of objects in $I_t$. 
We describe here our ConF technique in general, then specialize it to the object appearance property in sec.~\ref{sec:compsel} and to the object location property in sec.~\ref{sec:loc}.

\parskiny
\paragraph{ConF at training time} learns an ensemble of decision trees (forest) that operates on global image features $\phi(I)$. We construct each tree by recursively splitting the training set $\mathcal{T}$ at each node. We want the leaves of the trees to contain images
whose objects properties are compact according to some measure $c(\mathcal{T}_l)$, where $\mathcal{T}_l$ are the training images in leaf $l$. 
Each internal node $n$ contains a binary split function $f(\phi(I),\theta_n)$, where $\theta_n$ are its parameters.
Let $\mathcal{T}_n$ be the training images that reached node $n$, then $f(\phi(I),\theta_n)$ will split $\mathcal{T}_n$ into two subsets $\mathcal{T}_l$ and $\mathcal{T}_r$.
We use axis-aligned weak learners as $f$ \cite{criminisi2011}.
The split function $f(\phi(I),\theta_n)$ applies a threshold to one of the dimensions of image feature vector $\phi(I)$.
Following the extremely randomized forest approach~\cite{Moosman06}, for each node we randomly sample several thousand possible splits $\theta$ and choose one that maximizes the joint compactness:
\begin{equation}
\theta_n = \arg \max_{\theta} { c(\mathcal{T}_l) + c(\mathcal{T}_r)}
\end{equation}  
\[
\text{s.t.} \forall I \in \mathcal{T}_l, f(\phi(I),\theta) = 0, \forall I \in \mathcal{T}_r, f(\phi(I),\theta) = 1
\]
Compactness is defined as
\begin{equation}
c(\mathcal{T}) = \frac{1}{N^2} \frac{1}{\sigma^2 \sqrt{2 \pi}} \sum_{w_i \in \mathcal{T}} \sum_{w_j \in \{\mathcal{T} \backslash w_i\}} e^{-\frac{1}{2}\frac{D(w_i, w_j)^2}{\sigma^2}},
  \label{eq:component_compact}
\end{equation}
where $N$ is the number of ground-truth object bounding-boxes in set $\mathcal{T}$
and $D(w_i, w_j)$ is a distance measure between the properties of two object bounding-boxes $w_i$ and $w_j$. Note how the inner summation in eq.~(\ref{eq:component_compact}) is an estimation of the density of the distribution induced by all bounding-boxes in $\{\mathcal{T} \backslash w_i\}$, evaluated at $w_j$. This value is high if $w_j$ has other bounding-boxes nearby. The estimate is done with a Gaussian Kernel Density estimator~\cite{parzen62} (KDE).
We learn the standard deviation $\sigma$ {\em from the entire training set} once before we train the forest. This determines the scale of the problem, i.e. at which range of distances two bounding-boxes should be considered close. 
We compute $\sigma$ as follows. For each training bounding-box $w_i$ we compute its k-nearest neighbours in the whole training set and compute the standard deviation over them. Finally, we set $\sigma$ as the median of these standard deviations over all bounding-boxes. 

By employing different compactness measures $c$ we can use ConF to learn relations between different object properties and global image features. Later we show how to use it for selecting components relevant for a test image (sec.~\ref{sec:compsel}), for estimating likely object locations in a test image (sec.~\ref{sec:loc}).

\parskiny
\paragraph{ConF at test time} operates in two phases (see fig.~\ref{fig:highway} and \ref{fig:schema}). First, test image $I_t$ is passed through the forest, reaching a leaf in each tree. 
Thereby, each tree selects the subset of training images contained in that leaf.
We now accumulate these selections over all trees in the forest to form the score  $\eta(I_i,I_t)$ for $I_i \in \mathcal{T}$, which is the number of trees that have selected $I_i$.
We now construct the retrieval set $\mathcal{R}$ by selecting the $k$ most frequently selected training images. In our experiments $k=10$.

Note how the split function $f$ and the compactness measure $c$ operate in structurally different spaces. While $f$ operates on the global image features $\phi(I)$, $c$ is measuring the similarity of properties of {\em objects inside the images}.
In this fashion ConF learns the relation between the two. Importantly, $c$ is neither convex nor differentiable in $\phi(I)$ and we are only able to learn this inter-space relation thanks to the unique advantages offered by Random Forests.


\subsection{ConF for component selection}
\label{sec:compsel}

Here we assume that each component of an object detector has been previously trained from a visually compact set of object instances, and that we know the component id $\xi_j$ of each training instance $j$. In EE-SVMs each training instance leads to a unique component. In DPM the component id of a training instance can be inferred by the output of the training procedure~\cite{DPMrelease5}.
Based on this information, we train a ConF to select a small subset of components to run on a given test image $I_t$.

\parskiny
\paragraph{Training.}
To train ConF for components selection we define the distance $D(w_i,w_j)$ in eq.~(\ref{eq:component_compact}) as the L2 distance between the HOG descriptors of object bounding-boxes $w_i$ and $w_j$.

\parskiny
\paragraph{Test.}
We pass the test image $I_t$ through ConF obtaining a retrieval set $\mathcal{R}$.
We then estimate a posterior distribution $p(\xi_j | I_t)$ over detector components $\xi_j$ given the test image $I_t$.
As a training image $I$ might contain multiple instances from different components, each training image is `labelled' by a distribution over components $p(\xi_j | I)$.
We estimate the component distribution for the test image $I_t$ as the average over the training images in the retrieval set $\mathcal{R}$
\begin{equation}
p(\xi_j|I_t) =  \frac{1}{|\mathcal{R}|} \sum_{I \in \mathcal{R}} p(\xi_j|I)
\label{eq:post}
\end{equation} 
Based on this distribution, we can now select which components to run on $I_t$. We rank components by their probability and iteratively pick them until their combined probability mass exceeds a threshold $\gamma$.
This threshold controls a trade-off between running few components and getting high detection performance.
%
An interesting aspect of our formulation is that the number of selected components changes depending on the test image.
A test image with a characteristic appearance matching training images with a systematic recurrence of a few components will lead to a peaky $p(\xi_j|I_t)$.
In this case it is safe to run only a few components and we obtain a substantial speedup. 
On the other hand, if the ConF is uncertain about the contents of the test image,
then the entropy of $p(\xi_j|I_t)$ will be high, and many components will be selected. In the extreme case, for a very difficult test image, our procedure naturally degenerates to the default case of running all components.

\subsection{ConF for object location}
\label{sec:loc}

At test time, a typical detector scores hundreds of thousands of windows over the whole test image $I_t$, based on their appearance only.
We propose here to augment the detector's scores by adding knowledge about likely positions and scales of the object class, derived purely from the global appearance of $I_t$.

\parskiny
\paragraph{Training.}
We train two ConFs to predict likely object positions and scales, respectively. To do so, we employ a different measure of compactness, substituting the distance function between two windows in eq.~(\ref{eq:component_compact}) with $D_{\text{POS}}$ (or $D_{\text{SCALE}}$).
We define $D_{\text{POS}}(w_i, w_j)$ as the L2 distance between the centres of object bounding-boxes $w_i$ and $w_j$. We define $D_{\text{SCALE}}(w_i , w_j) = \max(\frac{H_i}{H_j},\frac{H_j}{H_i}) \cdot \max(\frac{W_i}{W_j},\frac{W_j}{W_i})$ as the difference in their scale ($W$ and $H$ refer to width and height).



\parskiny
\paragraph{Test.}
At test time, we first pass the test image $I_t$ through ConF obtaining a retrieval set $\mathcal{R}$, and then compute the following score for each window $w$ in the test image
\begin{equation}
  \frac{1}{N} \sum_{w_i \in \mathcal{R}} \frac{1}{\sigma^2 \sqrt{2 \pi}} e^{-\frac{1}{2}\frac{D(w,w_i)^2}{\sigma^2}}
  \label{eq:kde}
\end{equation}
where $N$ is the number of object instances in the retrieval set $\mathcal{R}$, and $D$ is either $D_{\text{POS}}$ or $D_{\text{SCALE}}$ . We learn $\sigma$ from the entire training set as in sec.~\ref{sec:acs}. This score captures how likely a window is to cover an object based on its location or scale. Finally, we linearly combine the location and scale scores with detector's score of a test window $w$. The usual non-maxima suppression stage follows.


\begin{table}
\resizebox{\columnwidth}{!}
{\footnotesize
\hfill{}
\begin{tabular}{ |l|c|c|c|c|c|c| }
\hline
\multirow{2}{*}{} & \multicolumn{3}{c|}{Car} & \multicolumn{3}{c|}{Horse}\\
\cline{2-7}
 & +Imgs & Objs & -Imgs & +Imgs & Objs & -Imgs\\
\hline
PASCAL12 \cite{pascal-voc-2012} & 1161 & 2017 & 1161 & 482 & 710 & 482\\
ImageNet \cite{Deng:CVPR2009} & 6383 & 7120 & 6383 & 4550 & 6631 & 4550\\
SUN2012 \cite{xiao10cvpr} & 828 & 1779 & 828 & - & - & -\\
Labelme \cite{russel:08:ijcv} & 6566 & 16743 & 6566 & - & - & -\\
UIUC \cite{Agarwal04} & 828 & 889 & 500 & - & - & -\\
PASCAL-10x \cite{zhu12bmvc} & - & - & - & 4065 & 6454 & 4065\\
\hline
Total & 15766 & 28548 & 15438 & 10107 & 13071 & 10107\\
\hline \hline
\textbf{Our training set} & 14125 & 25774 & 13830 & 9097 & 12407 & 9097\\
\textbf{Our test set} & 1641 & 2774 & 1608 & 1010 & 1388 & 1010\\
\hline 
\end{tabular}}
\hfill{}
\caption{\small \it Statistics of the large-scale dataset we assembled by combining images from existing source datasets. The column `+Imgs' reports the number of positive images per dataset; `Objs' is the total number of instances of the class in all positive images; `-Imgs' is the number of negative images we sampled.}
\figshrinky
\label{table:dataset}
\end{table}

\section{Multi-component detectors on large training sets} 
\label{sec:large_scale_MC}

In this section we study how multi-component detectors behave when trained on a large training set. We observe that it is necessary to increase the number of components as the size of the training set grows, so as to absorb the additional intra-class appearance variation. This motivates using ConF for component selection to speed-up the resulting large mixture models at test time.
We perform experiments (sec.~\ref{sec:largeData}) with two multi-component detectors (DPM~\cite{felzenszwalb10pami} and EE-SVM~\cite{MalisiewiczICCV11}, sec.~\ref{sec:MC_detectors}) on a large-scale dataset (sec.~\ref{sec:dataset}).
A related study was done by~\cite{zhu12bmvc}, using mainly a single component HOG detector (\cite{zhu12bmvc}, fig. 3-8). Their only experiment with a DPM~\cite{felzenszwalb10pami} is on a small training set of 900 faces (\cite{zhu12bmvc}, fig. 9-10). Hence, our study extends~\cite{zhu12bmvc} to large-scale training of DPMs and EE-SVMs.

\seckiny
\subsection{Multi-component detectors}
\label{sec:MC_detectors}

\paragraph{DPM\textnormal{~\cite{felzenszwalb10pami}}} represents an object class as a collection of parts arranged in a deformable configuration.
DPMs use a mixture of components, each specialized to an aspect of the training data to better capture the variation in appearance that the class exhibits.
Each component is trained on a subset of the training data with compact appearance, e.g. different viewpoints~\cite{felzenszwalb10pami} or subclasses~\cite{divvala2012eccv}. We use the publicly available implementation~\cite{DPMrelease5}.
 
\parskiny
\paragraph{EE-SVM\textnormal{~\cite{MalisiewiczICCV11}}} is a model composed of a separate linear SVM classifier for every training instance (exemplar). Each exemplar is represented by a rigid HOG template. The SVM is trained using the exemplar as the only positive against all negatives in the training set.
We refer to a single exemplar SVM as a component of the EE-SVM model, by analogy with DPM components.
At test time, each component is run on the image independently, producing many candidate detections. These are then filtered by non-maxima suppression in a final stage, where different components compete for the same image region. 
We use the publicly available implementation~\cite{malisiewicz2011eesvm}.
We assemble a large-scale dataset of two classes: car and horse (table \ref{table:dataset}). Below we discuss the car dataset in detail. The horse dataset was designed analogously.

\begin{figure*}[t]
\begin{subfigure}[b]{0.495\textwidth}
\includegraphics[width=\textwidth]{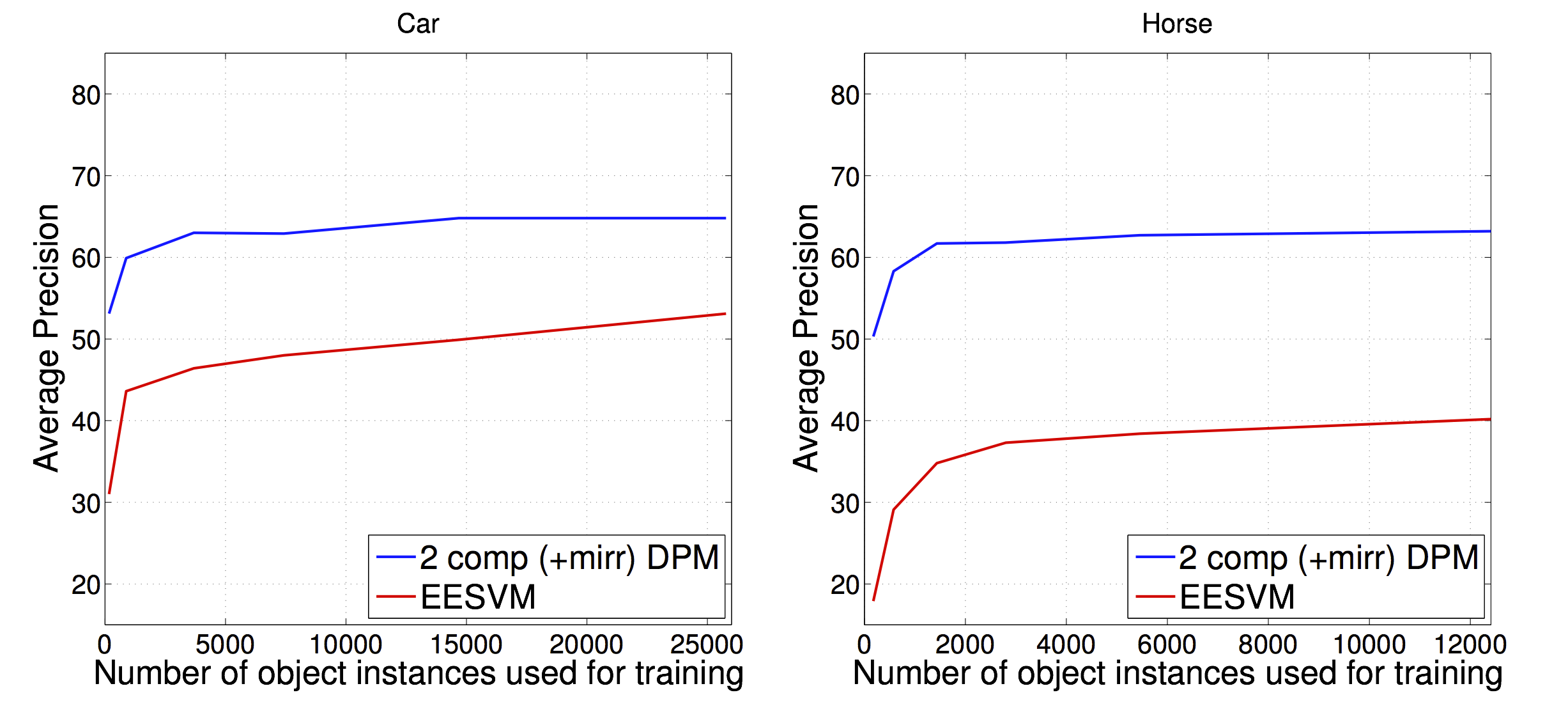}
\caption{\small \it Evolution of AP}
\label{fig:results_dg}
\end{subfigure}
\begin{subfigure}[b]{0.495\textwidth}
\includegraphics[width=\textwidth]{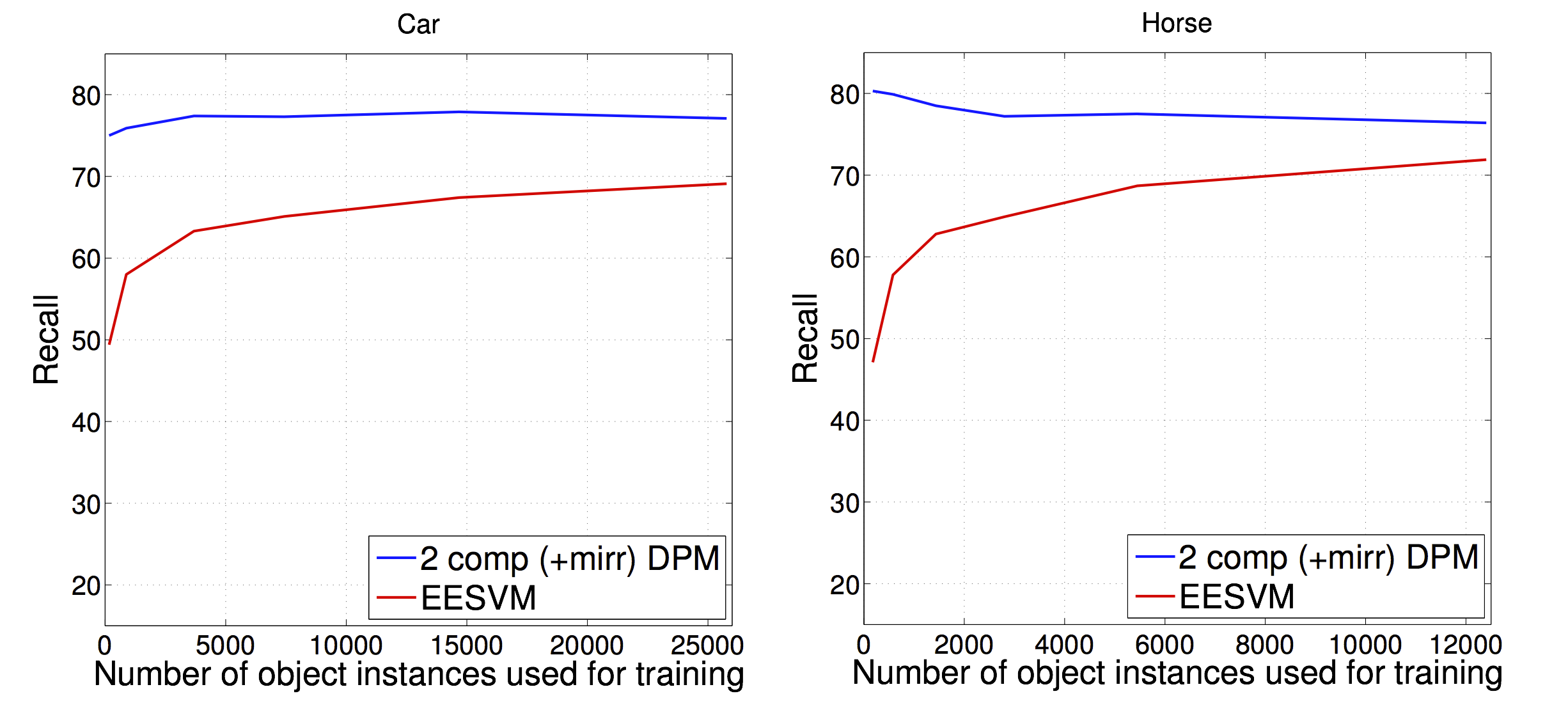}
\caption{\small \it Evolution of Recall} 
\label{fig:results_mc}
\end{subfigure}
\caption{\small \it  Results of experiments when training DPMs and EESVMs with increasing amounts of training data.}
\figshrinky
\label{fig:results_exp}
\end{figure*}

\subsection{Dataset}
\label{sec:dataset}
\vspace{3mm}
\parskiny
\paragraph{Source datasets.}
We combine 6 existing datasets:
PASCAL VOC 2012 \cite{pascal-voc-2012}, ImageNet \cite{Deng:CVPR2009}, LabelMe \cite{russel:08:ijcv}, SUN 2012 \cite{xiao10cvpr}, UIUC \cite{Agarwal04} and PASCAL-10x \cite{zhu12bmvc}.
%
%
PASCAL VOC 2012, PASCAL 10x, and ImageNet contain a variety of images, with both difficult, cluttered images and easier images with big centred cars.
UIUC has low resolution, gray-scale images of side-view cars.
LabelMe and Sun 2012 contain wide open street scenes with small cars.

\parskiny
\paragraph{Positive images and ground-truth annotations.}
We collected all images with bounding-box annotations on cars.
We took several steps to ensure a clean dataset.
First, we removed duplicate images, which were a few hundreds. Next, we removed incorrect bounding-boxes not covering cars or covering a car multiple times.
Finally, as some images have unannotated cars, we annotated all missing instances with bounding-boxes for our entire test set (e.g. images from ImageNet). This enables reliable performance measurements.

\parskiny
\paragraph{Negative images.}
We collect negative images from each dataset, so that it contributes an equal number of positive and negative images.
To ensure variation, we randomly sampled these negative images.

\parskiny
\paragraph{Train/test splits.}
We split the dataset into training (90\%) and test (10\%) sets. Then we split the training set further into 6 increasingly large subsets, which we use in sec.~\ref{sec:largeData} to study how the performance of the detectors evolves with increasing training data.
%
%
We ensure that each split contains images from all source datasets {\em in the same proportions} to avoid dataset bias issues~\cite{Torralba11}. These proportions correspond to the percentage of images that each source dataset contributes to the entire dataset (e.g. 8\% of all car images are from PASCAL VOC 2012, and 40\% from ImageNet).


\subsection{Experiments}
\label{sec:largeData}

We evaluate the performance of DPM~\cite{felzenszwalb10pami} and EE-SVM~\cite{MalisiewiczICCV11} as a function of the amount of training data and model capacity. 
First, we analyse how the performance on a fixed test set changes as the amount of training data increases. EE-SVM increase its capacity naturally with each positive example.
DPM instead needs its capacity controlled manually.
In the second experiment we increase the capacity of the DPM trained on the largest training set, hoping that it will help absorbing the training data better.

\parskiny
\paragraph{Increasing the amount of training data.}
For each class, we split the training set into 6 nested subsets, which contain 1\%, 5\%, 10\%, 25\%, 50\% and 100\% of the training images (sec. \ref{sec:dataset}). Each set is contained in all larger ones, and is composed of images from all source datasets in the same proportions.
Fig.~\ref{fig:results_exp} shows the average precision (AP) and recall on our test set. 

For DPM, we use 4 mixture components (2 and their mirrored versions). For both classes, increasing the amount of training data yields a modest improvement in AP.
%
Initially, performance increases roughly logarithmically in the number of training images, but eventually saturates. Surprisingly, this happens quite early.
A DPM car detector trained on the whole dataset only performs 1\% better than when trained on a third of the data. We observe similar behaviour on horses.
Recall, on the other hand, saturates almost immediately for cars
and even decreases for horses as the training set grows.


The EE-SVM detector demonstrate continuous, non-saturating growth in both AP and recall as the training data increases. The growth in recall is particularly strong, as it improves by more than 20\% for both classes after seeing the full training set.

\begin{figure}[t]
\includegraphics[width=1.05\textwidth]{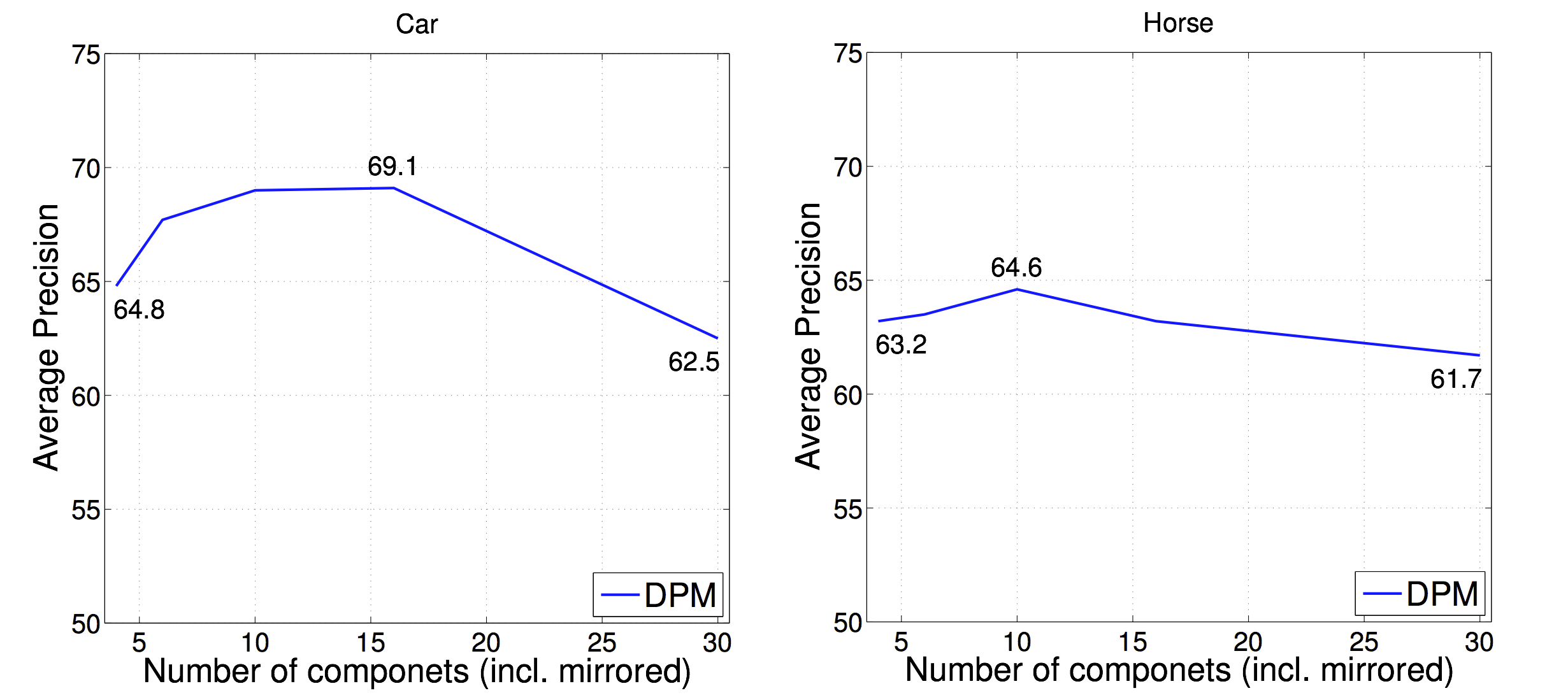}{\caption{\small \it Evolution of AP when increasing the number of DPM components. The performance increases steadily, but eventually it starts overfitting.}
\figshrinky
\label{fig:dpm_capacity_grow}}
\end{figure}

\parskiny
\paragraph{Increasing DPM capacity.}
In fig.~\ref{fig:dpm_capacity_grow} we help DPM to better absorb the training data by progressively increasing its number of components, while keeping the training set fixed to  the largest one.
For both classes, performance increases steadily as the number of components grows from 4 (2+2 mirrored) to 16 (8+8) for cars and 10 (5+5) for horses. After that the model starts to overfit: performance decreases and eventually (30 components) drops below the performance of the smallest model (4 components).  For cars, each component clearly represents a different aspect (mostly view-point). 

This overfitting behaviour, even on such a large training set is an interesting finding. The practical implication is that the DPM user has to be careful and manually control the capacity to obtain the best performance.
In contrast, EE-SVM does not overfit even when training $>20000$ components.
Yet, while in terms of growth EE-SVM behaves very well, we note that its absolute detection performance is lower than DPMs.


\subsection{Conclusions}

In general, both DPM and EE-SVM do benefit from large training sets. The key to continued growth for both methods is control of capacity. EE-SVM automatically increases its capacity, by adding a new component with each training sample.
For DPM, the best results are achieved through selecting the right capacity manually, which also corresponds to a rather large number of components (compared to the 2+2 traditionally used on Pascal VOC~\cite{pascal-voc-2012}). Using such large models comes at the cost of longer runtime, as all components have to be applied to a test image. This is especially problematic for EE-SVMs, as running ten thousand components takes about 10 minutes for a single image.
In the next section we demonstrate how ConF substantially reduces this computational burden by selecting only a small subset of components most relevant to a particular test image and run only those.


%
\begin{figure*}[t]
\begin{subfigure}[b]{0.495\textwidth}
\includegraphics[width=\textwidth]{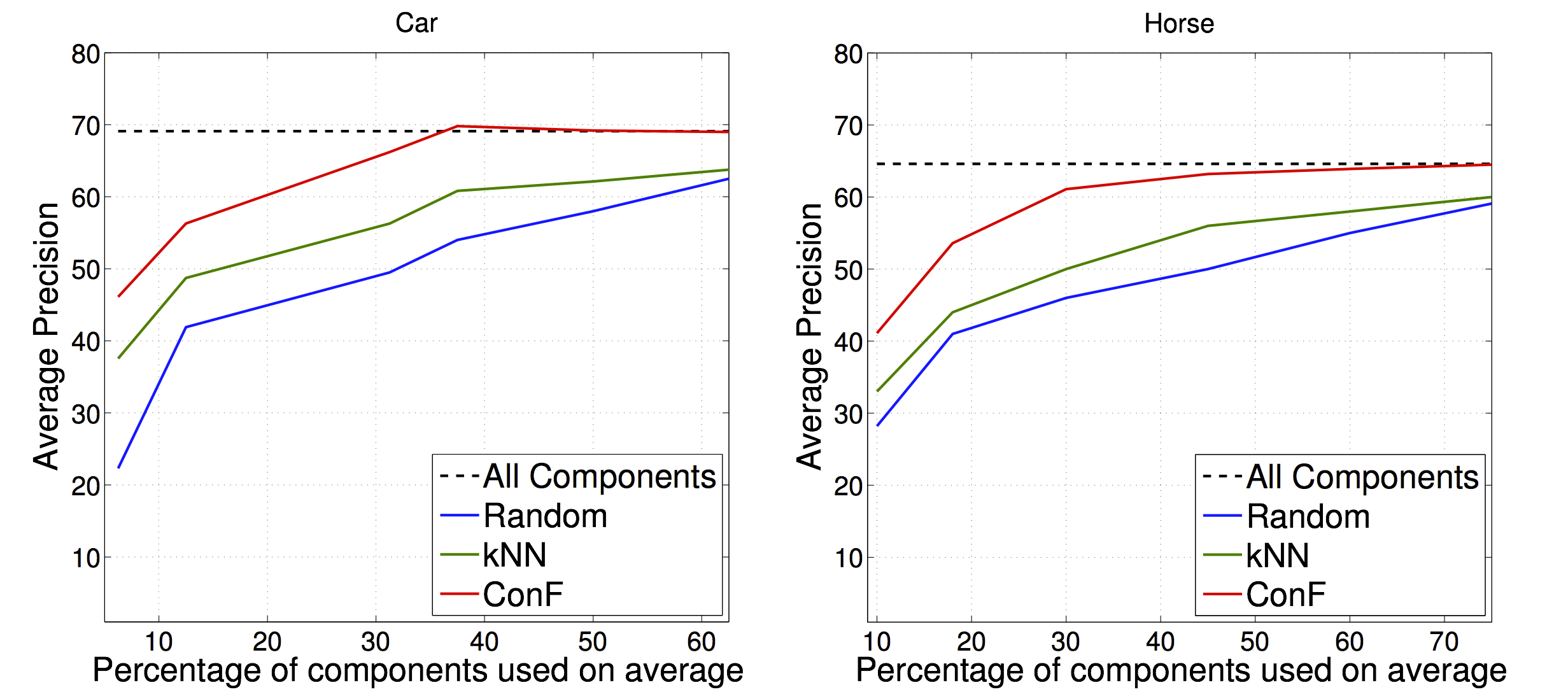}
\caption{DPM} 
\label{fig:c1}
\end{subfigure} 
\begin{subfigure}[b]{0.495\textwidth}
\includegraphics[width=\textwidth]{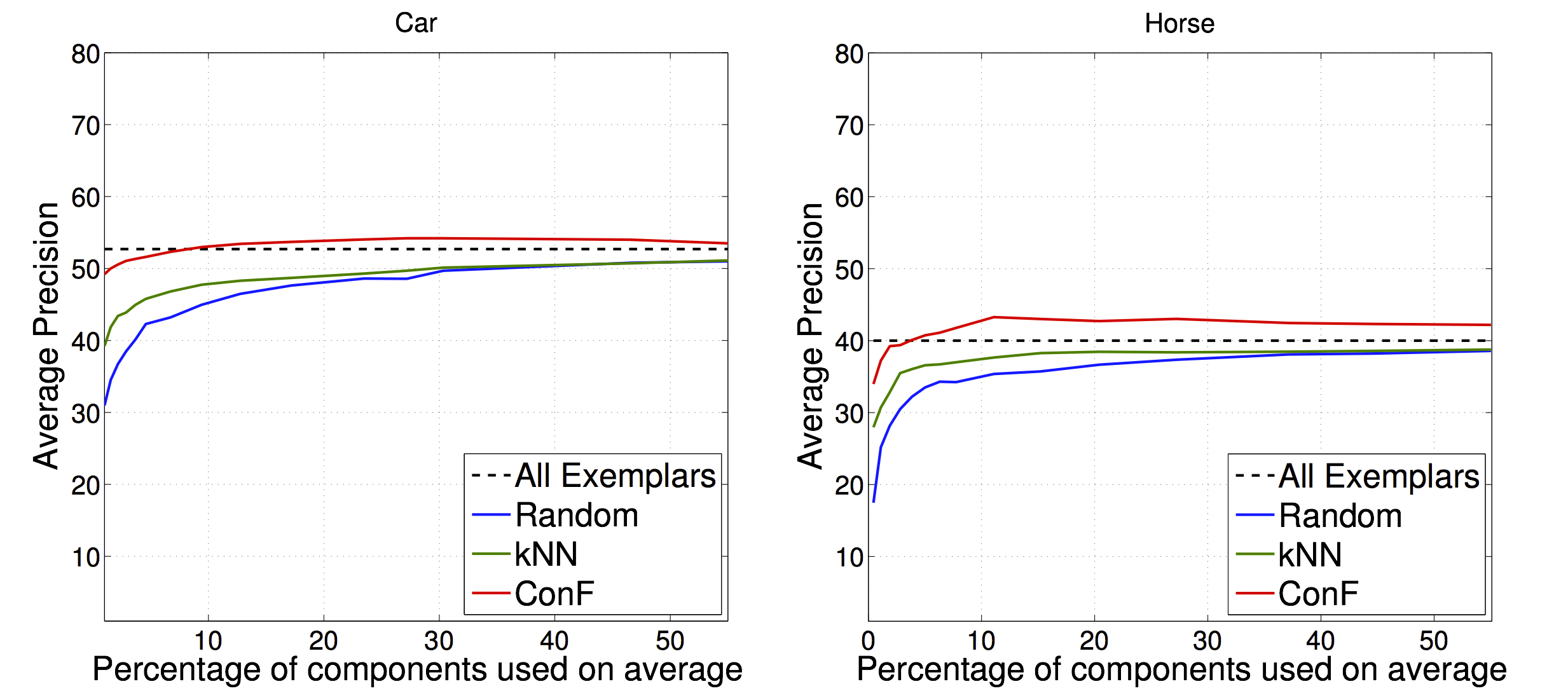}
\caption{EESVM} 
\label{fig:c2}
\end{subfigure}
\caption{\small \it  Results of applying ConF for the automatic component selection. The points on the plot correspond to different choices for the threshold $\gamma$ (sec.~\ref{sec:compsel}). The horizontal axis corresponds to the average amount of components used. The vertical axis corresponds to the AP of the detector using components selected by ConF.}
\figshrinky
\label{fig:exp_comp_sel}
\end{figure*}

\begin{figure*}
\includegraphics[width=\textwidth]{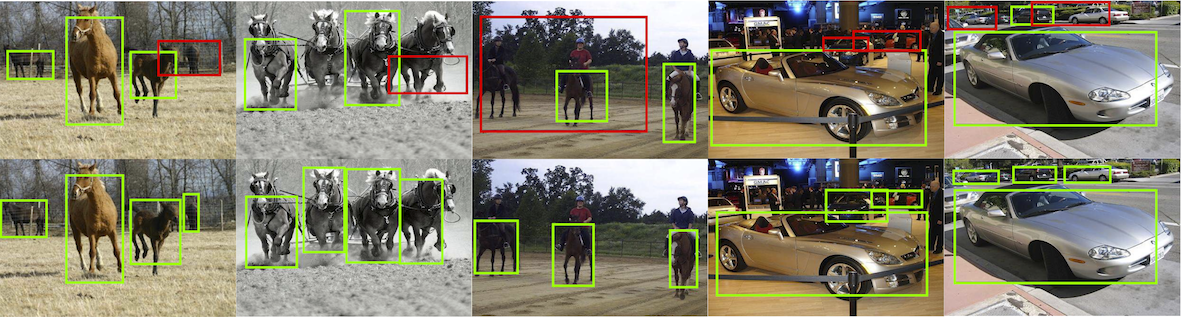}
\caption{\it \small Detection obtained before (top row) and after (bottom row) applying ConF for component selection. Green bounding-boxes highlight correct detections, while red ones show false positives.} 
\label{fig:res_com_sel}
\figshrinky
\end{figure*}

\begin{figure*}
\includegraphics[width=\textwidth]{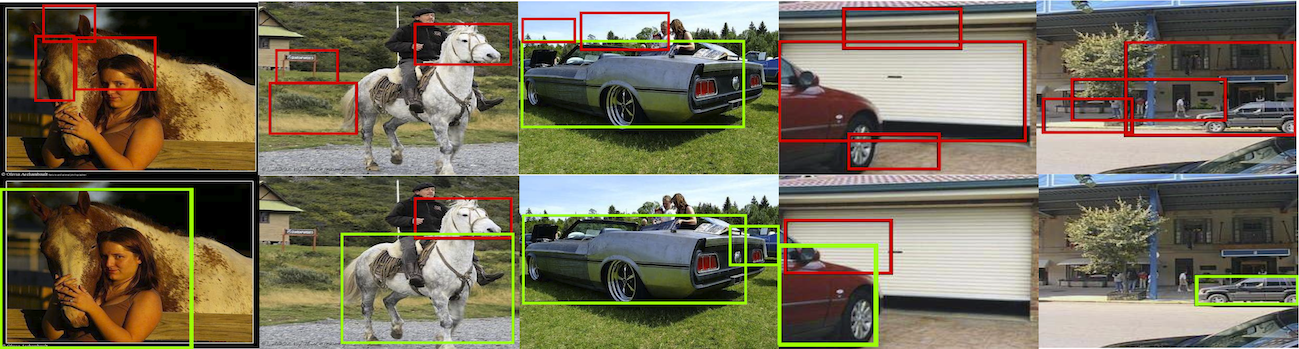}
\caption{\it \small Detection obtained before (top row) and after (bottom row) applying ConF as location model. Green bounding-boxes highlight correct detections, while red ones show false positives. } 
\figshrinky
\label{fig:res_loc_models}
\end{figure*}

\section{Experiments with ConF} 
\label{sec:exp}

In this section we evaluate the performance of ConF for component selection and location estimation. As global image descriptors $\phi(I)$ we extract SURF \cite{Bay08}, LAB and SIFT \cite{LoweIJCV04} descriptors on a dense grid at multiple scales. 
For each feature type we train a class-specific codebook of 1000 visual words and construct a 2-level spatial pyramid~\cite{Lazebnik06fixed}. Additionally, we also extract a GIST \cite{Oliva01} descriptor for the image. Overall, we train ConF on a 16000 dimensional feature space, using 750 trees for each task.

\parskiny
\paragraph{Quality of retrieval sets.}
We quantify how similar object bounding-boxes from a test image $I_t$ are to those in the retrieval set $\mathcal{R}$ returned by the method as follows

\begin{equation}
 \frac{1}{Z \sigma^2 \sqrt{2 \pi}} \sum_{w_i \in I_t} \sum_{w_j \in \mathcal{R}(I_t)} e^{-\frac{1}{2}\frac{D(w_i, w_j)^2}{\sigma^2}},
  \label{eq:direct_measurement}
\end{equation}
where $Z$ is number of pairs of bounding-boxes in $I_t$ and $\mathcal{R}$.
The distance $D$ and standard deviation $\sigma$ vary depending on the property (appearance, position, scale) as defined in sec.~\ref{sec:compsel},~\ref{sec:loc}.

Table~\ref{table:kde} show results averaged over the test set, higher values are better. 
As a baseline, we return the whole training set as the retrieval set. This leads to a generic prior on image properties, independent of the test image.
Moreover, we compare to the traditional way of building retrieval sets by k-nearest neighbours (kNN)~\cite{liu09cvpr,russell07nips,Torralba03, tighe10eccv,tighe13cvpr}, defined on the same features as ConF.
Both kNN and ConF greatly outperform the baseline, proving they return meaningful retrieval sets. This confirms the observation that the global image appearance conveys information about the objects properties inside the images.
Moreover, ConF returns better retrieval sets than kNN across all object properties and retrieval set sizes evaluated.

\begin{table*}[t]
\small \centering
{
\begin{tabular}{ |l||c|c c|c c||c|c c|c c| } 
\hline
\multirow{3}{*}{\backslashbox{Obj property}{Data}} & \multicolumn{5}{c||}{\it Car} & \multicolumn{5}{c|}{\it Horse} \\\cline{2-11} 
& All train  & \multicolumn{2}{c|}{NN} & \multicolumn{2}{c||}{ConF} & All train  & \multicolumn{2}{c|}{NN} & \multicolumn{2}{c|}{ConF} \\  
&  data &  1 & 10 & 1 & 10 & data &  1 & 10 & 1 & 10 \\\hline
Appearance & 0.01 & 0.10 & \multicolumn{1}{|c|}{0.03} & \multicolumn{1}{c|}{\bf 0.12} & 0.09 & 0.03 & \multicolumn{1}{c|}{0.04} & 0.04 & \multicolumn{1}{c|}{\bf 0.06} & 0.05\\
Position & 0.36 & 1.30 & \multicolumn{1}{|c|}{0.88} & \multicolumn{1}{c|}{\bf 1.84} & 1.42 & 0.37 & \multicolumn{1}{c|}{0.50} & 0.53 & \multicolumn{1}{c|}{\bf 0.57} & 0.57\\
Scale & 0.03 & 0.05 & \multicolumn{1}{|c|}{0.04} & \multicolumn{1}{c|}{ \bf 0.08} & 0.07 & 0.06 & \multicolumn{1}{c|}{0.08} & 0.08 & \multicolumn{1}{c|}{\bf 0.09} & 0.08\\ 
\hline
\end{tabular}}
\caption{\small \it Evaluation of the quality of retrieval sets for predicting object properties. Each entry represents the average density of the retrieval set evaluated at the objects properties in the test images. We consider two sizes for the retrieval set $|\mathcal{R}| = 1$ and $|\mathcal{R}| = 10$. }
\label{table:kde}
\end{table*}

\paragraph{Automatic component selection.}
We now use ConF to select object detector components relevant for a given test image. We use the best performing settings from sec. \ref{sec:largeData}, i.e. 16 (10) component DPM for cars (horses) and {\em very large} EE-SVMs using all training exemplars, i.e. 25774 for cars and 12407 for horses.

Fig.~\ref{fig:exp_comp_sel} shows the evolution of AP while increasing the percentage of components used (higher is better). 
We compare to building retrieval sets by kNN, and to a baseline which randomly selects components without looking at the test image.
ConF outperforms the baseline and kNN for both object classes, for both detection models, and over the whole range of the plots. By employing ConF, we closely match the performance of the full DPM model by running roughly half of the components. We match the performance of a full EE-SVM when running less than 10\% of the components.
Even in the extreme case of running just {\em one} EE-SVM component, the AP is about 90\% of that of the full model.
Interestingly, for EE-SVM on the horse class, ConF \emph{improves AP by 3\%} over the full ensemble using all components, when running \emph{10$\times$ fewer components}. 
There is also a minor improvement in AP for EE-SVM on the car class, when running half of the components. The AP improvement comes from dropping some components that lead to false positives.

These experiments demonstrate the ability of ConF to select relevant components given just global image appearance. This makes EE-SVMs practical even when trained from large sets with tens of thousands of exemplars (the average runtime for a test image decreases from 10 minutes to 1 minute on our machine with 4 i5-core 3.10GHz processor and 16 GB memory). Fig.~\ref{fig:res_com_sel} shows some example results.

\begin{table}[t]
\small \centering
\resizebox{0.7\columnwidth}{!}
{
\begin{tabular}{|c|c|c|c|c|c|}
\hline 
\multicolumn{6}{|c|}{\it Car}\tabularnewline
\multicolumn{3}{|c|}{DPM} & \multicolumn{3}{c|}{EE-SVM} \tabularnewline
\hline 
None & NN & ConF & None & NN & ConF \tabularnewline
69.1 & 0 & +2.1 & 52.7 & 0 & +2.3 \tabularnewline
\hline 
\hline
\multicolumn{6}{|c|}{\it Horse} \tabularnewline
\multicolumn{3}{|c|}{DPM} & \multicolumn{3}{c|}{EE-SVM}\tabularnewline
\hline
None & NN & ConF & None & NN & ConF\tabularnewline
64.6 & 0 & +0.7 & 40 & 0 & +1.1\tabularnewline
\hline
\end{tabular}
\caption{\label{table:locandscale}\small \it The results of augmenting the detector score with the location model derived by ConF (sec.~\ref{sec:loc}) and NN compare to not using location model at all (None).}}
\end{table}

\paragraph{Object locations.}
Here we demonstrate how ConF trained to estimate the location of objects from global image features can improve detection performance by downgrading the score of false positives at unlikely locations. As tab.~\ref{table:locandscale} shows, this improves AP for both classes and both detectors
(+2\% for cars and +1\% for horses). Instead, kNN does not bring any improvement, further confirming that ConF returns better retrieval sets. Fig.~\ref{fig:res_loc_models} shows example results.
%

\paragraph{Computational and memory efficiency.}
ConF does not only offer better performance than kNN, but is also more memory and computationally efficient.
In terms of computation, kNN requires a number distance computations linear in the number of training images, where ConF requires only a logarithmic number of threshold operations. 

In terms of memory, kNN stores all feature vectors of all images in the training set. For cars, this amounts to 1.68 GB.
For each internal node ConF stores a threshold, a feature id and the ids of its children, amounting to 16 bytes. The leaves store the indices of the training images they contain, for a total of exactly the number of training images $\times 2$ bytes overall (per tree). For  cars, there are $<900$ internal nodes on average per tree. As we store 750 trees per class, the grand total is only 27 MB ({\em $60\times$} less than kNN).

\paragraph{Conclusions.}
We propose a novel method --- ConF, which learns the relation between the global image appearance and properties of objects in the image. We show how ConF can be employed to dynamically select a small number of components for a test image. This improves speed and, sometimes, even detection performance. We have also shown how to use ConF to predict likely object location to reduce false positive rates.

\newpage
{\small
\bibliographystyle{ieeetr}
\bibliography{../../bibtex/shortstrings,../../bibtex/vggroup,../../bibtex/calvin}
}
\end{document}